\documentclass{article}

\PassOptionsToPackage{square,numbers}{natbib}

     \usepackage[preprint]{neurips_2021}

\usepackage[utf8]{inputenc} 
\usepackage[T1]{fontenc}    
\usepackage{hyperref}       
\usepackage{url}            
\usepackage{booktabs}       
\usepackage{amsfonts}       
\usepackage{nicefrac}       
\usepackage{microtype}      
\usepackage{xcolor}         

\usepackage[nottoc]{tocbibind} 

\usepackage{amsmath,amsthm}
\usepackage{amsfonts}
\usepackage{amssymb}
\usepackage{graphicx}
\usepackage[ruled,linesnumbered,lined,commentsnumbered]{algorithm2e}
\usepackage{multirow}
\usepackage{wrapfig}

\newcommand{\Gc}{\mathcal{G}}
\newcommand{\Dc}{\mathcal{D}}

\newcommand{\Xc}{\mathcal{X}}
\newcommand{\Yc}{\mathcal{Y}}
\newcommand{\Zc}{\mathcal{Z}}

\title{Federated CycleGAN for Privacy-Preserving Image-to-Image Translation}

\author{%
  Joonyoung Song \quad \quad Jong Chul Ye \\
  Department of Bio and Brain Engineering\\
  Korea Advanced Institute of Science and Technology (KAIST), Daejeon, Korea \\
  \texttt{\{songjy18, jong.ye\}@kaist.ac.kr} \\
}

\begin{document}

\maketitle

\begin{abstract}
  Unsupervised image-to-image translation methods such as CycleGAN
    learn to convert  images from one domain to another using unpaired training data sets from different domains.
Unfortunately, these approaches still require centrally collected unpaired records, potentially violating privacy and security issues.
Although the recent federated learning (FL) allows a neural network to be trained without data exchange, the basic assumption of the FL is that all clients have their own training data from a similar domain, which is different from our  image-to-image translation scenario in which each client has images from its unique domain
and the goal is to learn image translation between different domains without accessing the target domain data.
To address this, here  we propose a novel federated CycleGAN architecture that can learn  image translation
in an unsupervised manner while maintaining the data privacy. Specifically,
our approach arises from a novel observation that CycleGAN loss can be decomposed into 
 the sum of client specific local objectives that can be  
 evaluated using only their data. 
 This local objective decomposition allows multiple clients to participate in federated CycleGAN training without sacrificing performance.
Furthermore, our method employs  novel
switchable generator and discriminator architecture using Adaptive Instance Normalization (AdaIN) that significantly
reduces the  band-width requirement of the federated learning.
  Our experimental results on various unsupervised image translation tasks
  show that our federated CycleGAN provides comparable performance 
  compared to the non-federated counterpart.
\end{abstract}

\section{Introduction}

  Unsupervised image-to-image (I2I) translation is to learn  image conversion from one domain to another using unpaired training data sets from two domains. Recent works \citep{zhu2017unpaired, liu2017unsupervised, yi2017dualgan, zhu2017toward, huang2018multimodal, lee2018diverse, liu2019few, Kim2020U-GAT-IT} have shown impressive results using a generative adversarial network (GAN) framework \citep{goodfellow2014generative} and their practical applications in various areas such as computer vision \citep{yuan2018unsupervised, lu2019unsupervised, du2020learning}, medical imaging \citep{kang2019cycle, oh2020unpaired, kim2021cyclemorph}, and remote sensing \citep{song2020unsupervised, luppino2021deep}, etc.
  However, most image-to-image translation methods such as cycleGAN \citep{zhu2017unpaired} still require centrally collected unpaired datasets that are often difficult to obtain due to privacy and security issues.
  
  \begin{figure}[!t]
\centering
\includegraphics[width=1\linewidth]{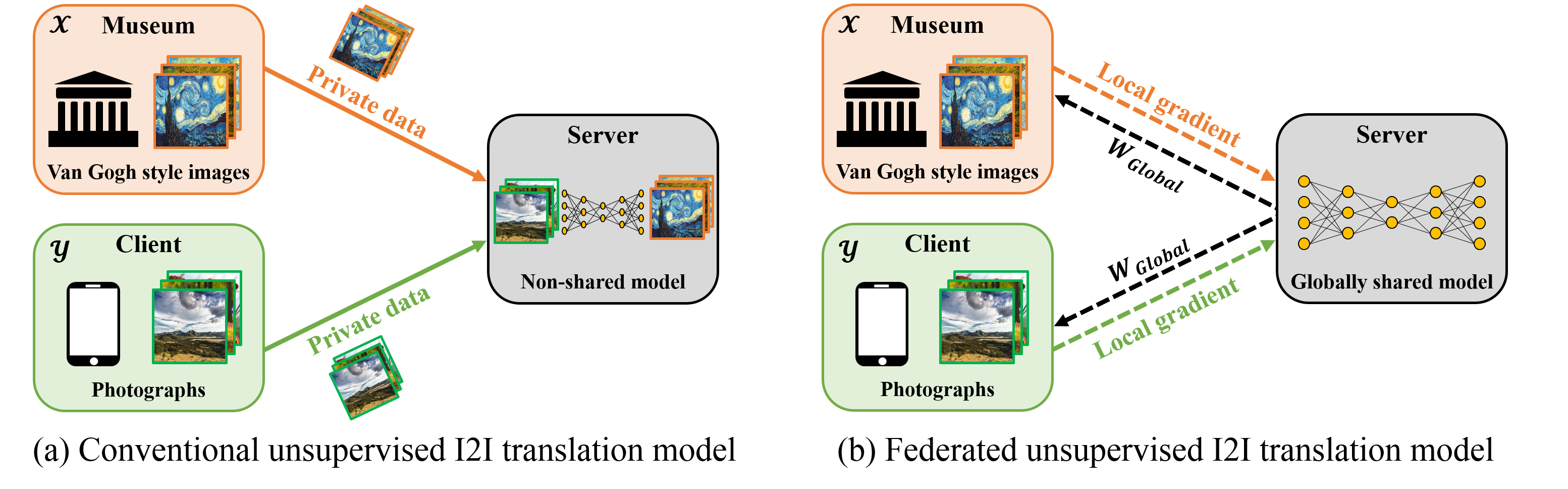}
\vspace{-0.5cm}
\caption{{Unsupervised image-to-image (I2I) translation using (a)  centrally collected data 
and (b) distributed data using federated learning. Federated learning trains the model without sharing private data.}}
\label{fig:concept}
\end{figure}

{For instance, suppose one wants to train a neural network model that can transfer photos to Van Gogh style images, where one client has only photographs while another client (eg. museum) has digital copies of artwork  by Van Gogh. 
However, conventional image translation approaches  require centrally collected photos and artwork  as shown Fig.~\ref{fig:concept}(a), which requires copyright of artwork
and is also vulnerable to data security bleaching even under the copyright agreement. 
  }
  
  Recently, federated learning (FL) \citep{mcmahan2017communication} has drawn a lot of interest because it ensures data privacy by not sharing private data. In federated learning, a central server sends parameters of the global model to multiple clients. Each client trains the local model using its own data and sends current updates to the server. Finally, the server aggregates local updates to train the global model. FedAvg \citep{mcmahan2017communication} is a representative algorithm for training a global model by averaging local updates from clients and is extended to other variants \citep{NIPS2017_6211080f, pmlr-v97-yurochkin19a, MLSYS2020_38af8613,  Wang2020Federated}. 
{Moreover, peer-to-peer direct communication protocol without central server has been also studied \cite{cola2018nips}.}

\begin{figure}[!t]
\centering
\includegraphics[width=1\linewidth]{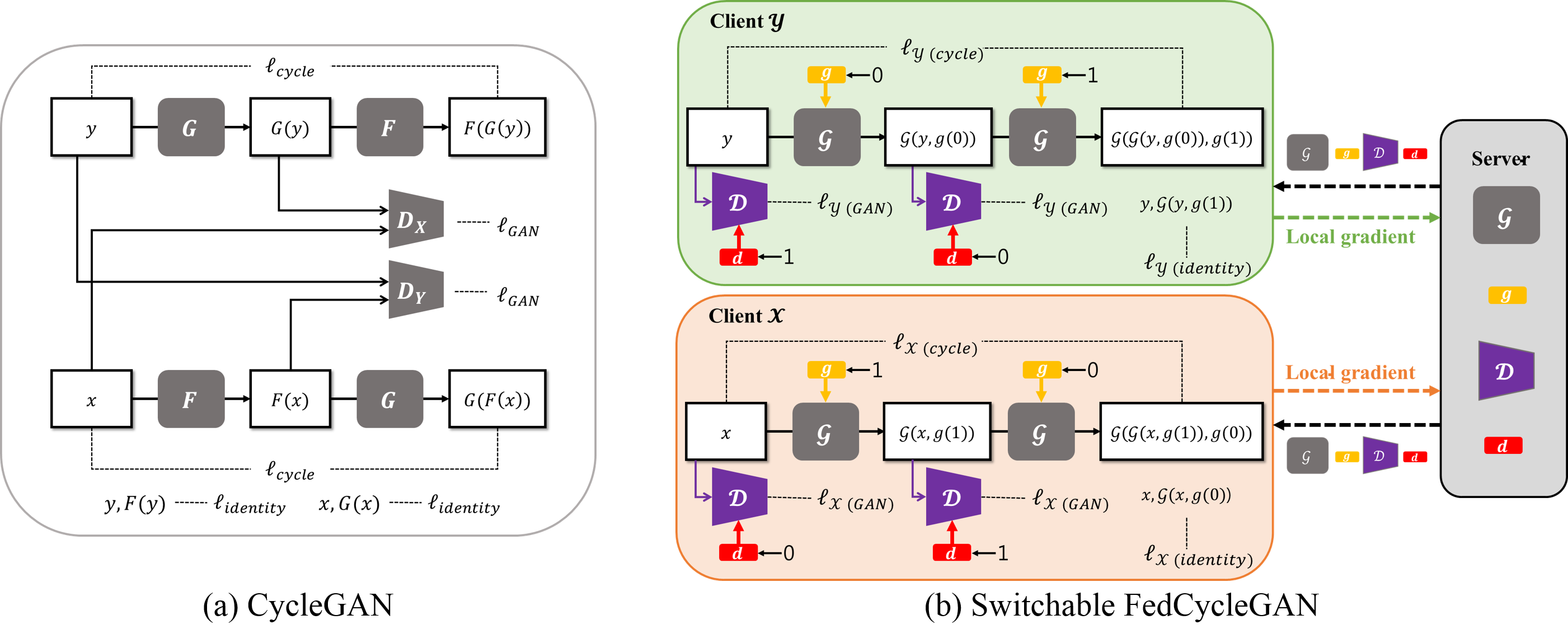}
\vspace{-0.7cm}
\caption{{Architectures of (a) CycleGAN and (b) our switchable FedCycleGAN.}}
\label{fig:FedCycleGAN}
\end{figure}

Recently, several works \citep{Augenstein2020Generative, neurips20chen} successfully trained  GANs in a federated learning scenario.
  Unfortunately, the application of federated learning to unsupervised image-to-image translations using CycleGAN is still an open problem. This is because that training image-to-image translations between two domains requires access to both domain data.  
  Specifically, as shown in Fig.~\ref{fig:FedCycleGAN}(a), to translate images from a domain $\Yc$ to a different domain $\Xc$, the generator $G:\Yc\mapsto \Xc$ converts images from the domain $\Yc$ to the domain $\Xc$ so that the discriminator $D_X$ cannot distinguish them with real samples from the domain $\Xc$. 
  To train the discriminator $D_X$, it requires both real samples from the domain $\Xc$ and fake samples synthesized by the generator $G$ using images from the $\Yc$ domain. However, in a federated learning setting in which one client (domain $\Yc$) is not allowed to access data from another client (domain $\Xc$) and a central server also never accesses data, the discriminator $D_X$ of the client (domain $\Yc$) cannot be trained because it cannot access real samples from the domain $\Xc$.


To address this problem, here  we propose a novel federated CycleGAN (FedCycleGAN), which can be trained without sharing local client data.
FedCycleGAN is possible thanks to the following key innovation.
Specifically, in contrast to the conventional FL where the same form of the total loss is used across clients,
our FedCycleGAN decomposes the total CycleGAN loss  into the sum of the
domain specific local objective of each client that can be computed using only their data.
Accordingly, the clients transmit the gradients of their own objective functions so that the server
then sums them up to compute the gradient of the total CycleGAN loss.
Accordingly,  the server trains the globally shared model using local gradients from clients without sharing private data and therefore protects privacy (see Fig.~\ref{fig:concept} (b)).
The proposed local objective decomposition is very scalable in the sense that any number of clients
can participate in federated CycleGAN training. {Specifically, as shown in Fig.~\ref{fig:protocol}}, each client
transmits local gradient of its domain-specific objective functions, which are then averaged
at the server to compute the global gradients.

Yet another innovation is the switchable architecture that significantly 
reduces the transmission bandwidth with negligible performance degradation. Specifically, inspired by AdaIN-based switchable CycleGAN \citep{gu2021adain, yang2021continuous},
our framework use  switchable architecture for generator and discriminator using adaptive instance normalization (AdaIN) \citep{huang2017arbitrary}
to reduce the transmission overhead.
Thus, we only need to transmit
gradients from  a single pair of generator and discriminator  instead of  two.
This significantly reduces the bandwidth requirement.
   The main contribution of this paper can be therefore summarized as following: 
   \begin{enumerate}
   \item In contrast to the conventional federated learning that uses the same loss function across all clients, in our FedCycleGAN
  clients use their own domain-specific loss functions to compute the local gradient, which are summed at the server to
   compute the global gradient.  In fact,     our experimental results demonstrate that our FedCycleGAN can show comparable and even better performance
    compared to  non-federated vanilla CycleGAN thanks to the exact local objective decomposition.
   \item Using AdaIN based switching scheme, the transmission overhead for federated learning can be significantly
   reduced, which makes the federated CycleGAN more practical.
   \end{enumerate}

\section{Related work}

\paragraph{Federated learning (FL)}
FL is a decentralized learning approach where local clients train their local models without transmitting data to a central server, and the global model is updated by aggregating local updates from clients \citep{mcmahan2017communication}. To aggregate local computations from clients, FedAvg \cite{mcmahan2017communication} updates a global model by averaging local updates. \citep{NIPS2017_6211080f} shows that federated learning can be applied to multi-task learning. \citep{pmlr-v97-yurochkin19a} propose a probabilistic federated learning approach based on Bayesian nonparametric framework. FedProx \citep{MLSYS2020_38af8613} is proposed to address the heterogeneous nature of federated learning. FedMA \citep{Wang2020Federated} improves performance by matching and averaging local updates. 
   
By extending the scope of classical FLs, several recent approaches \citep{Augenstein2020Generative, neurips20chen} have successfully integrated federated learning into GAN framework.
{For example, \citep{Augenstein2020Generative} uses DP-FedAvg-GAN to train GAN with differential privacy guarantees for an image synthesis. 
Specifically, as shown in Fig.~\ref{fig:comp}(a), a server in DP-FedAvg-GAN has a shared generator $G$ and a discriminator $D_X$ that are passed to clients. Each client updates a local discriminator using its own data ($x \sim P_\Xc$) and fake images ($G(z)$) and then sends a local update to the server. Finally, the server updates the global discriminator using local updates from clients, and then trains the global generator. This process is repeated until the convergence is achieved.
   GS-WGAN \citep{neurips20chen} employs a gradient-sanitized Wasserstein GAN approach to preserve privacy. As shown in Fig.~\ref{fig:comp}(b), a server has a centralized generator and transmits only synthesized images ($G(z)$). Each client calculates a local update using its own discriminator ($D_X$) and data ($x \sim P_\Xc$), and then transmit a sanitized local gradient to the server. The server trains the generator using local gradients from clients. 

\begin{figure}[!t]
\centering
\includegraphics[width=0.9\linewidth]{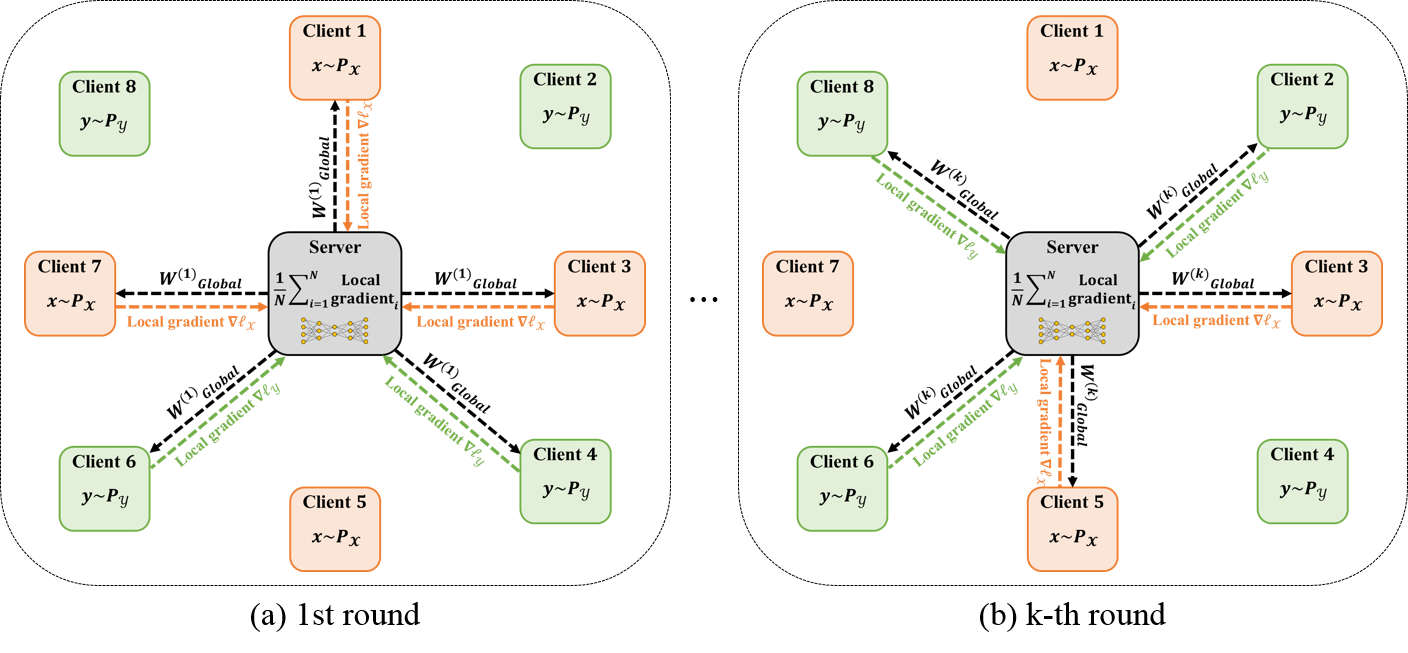}
\vspace{-0.5cm}
\caption{{FedCycleGAN training for multiple clients. At each round, a server selects $N$ clients randomly and local stochastic gradients from $N$ clients are averaged to train networks.}}
\label{fig:protocol}
\end{figure}

\paragraph{CycleGAN for unsupervised image-to-image translation}

   The goal of an unsupervised image-to-image translation is to learn how to translate a image from one domain ($\Yc$) to a corresponding output image in another domain ($\Xc$). Fig.~\ref{fig:FedCycleGAN} (a) shows the architecture of CycleGAN for this purpose.
   Suppose that $P_\Xc$ is a probability distribution of $\Xc$, and $P_\Yc$ is that of $\Yc$. $x$ and $y$ are images from $\Xc$ and $\Yc$, respectively. The generator $G:\Yc\mapsto \Xc$  translates an image from $\Yc$ to an output image in $\Xc$. The discriminator $D_X$ distinguishes real samples in $\Xc$ and fake samples that are generated by $G$ using samples in $\Yc$. Similarly, $F:\Xc\mapsto \Yc$ is the generator that translates an image in $\Xc$ into a corresponding output in $\Yc$. The discriminator $D_Y$  distinguishes real images in $\Yc$ from fake images that are made by $F$ using images in $\Xc$.
   
   In CycleGAN, the total loss function constitutes of adversarial loss and cycle-consistency loss. 
 The adversarial loss for the generator $G$ and the discriminator $D_X$ is given by
\begin{align}
\begin{split}
\ell_{GAN}(G,D_X) & =\mathbb{E}_{x \sim P_\Xc}[\log D_X(x)] + \mathbb{E}_{y \sim P_\Yc}[\log (1-D_X(G(y)))].
\label{eqn:cyclegan_gan_loss}
\end{split}
\end{align}
 whereas the adversarial loss $\ell_{GAN}(F,D_Y)$ for $F$ and $D_Y$ is: 
\begin{align}
\begin{split}
\ell_{GAN}(F,D_Y) & =\mathbb{E}_{y \sim P_\Yc}[\log D_Y(y)] + \mathbb{E}_{x \sim P_\Xc}[\log (1-D_Y(F(x)))].
\label{eqn:cyclegan_gan_loss2}
\end{split}
\end{align}
   As the adversarial loss does not guarantee the one-to-one mapping between an input and an output, cycle-consistency loss is necessary,
   which is 
   formulated  as follows:
\begin{align}
\begin{split}
\ell_{cycle}(G,F) & =\mathbb{E}_{y \sim P_\Yc}[||F(G(y))-y||_1] + \mathbb{E}_{x \sim P_\Xc}[||G(F(x))-x||_1],
\label{eqn:cyclegan_cycle_loss}
\end{split}
\end{align}
%
%
%
%
%
%
%

 The total loss with the adversarial loss and the cycle-consistency loss is defined as follows:
\begin{align}
\begin{split}
\ell_{CycleGAN}(G,F,D_X,D_Y) & = \ell_{GAN}(G,D_X)  + \ell_{GAN}(F,D_Y) + \lambda \ell_{cycle}(G,F) 
\label{eqn:cyclegan_overall_loss}
\end{split}
\end{align}
where $\lambda$  controls the weights of each component in the total loss. To train an unsupervised image-to-image translation in the CycleGAN framework, the following problem needs be solved:
\begin{align}
\min_{G,F}\max_{D_X,D_Y}\ell_{CycleGAN}(G,F,D_X,D_Y).
\label{eqn:cyclegan_min-max_overall_loss}
\end{align}
By solving the minmax problems, the generators $G$ and $F$ aim to generate realistic images in order to deceive the discriminator $D_X$ and $D_Y$, respectively, and the discriminators try to distinguish real samples  and fake samples  that are synthesized by the generator. 
Additionally, we often use an identity loss $\ell_{identity}$ in \eqref{eqn:cyclegan_overall_loss} to
 enforce a generator to retain input that is from the target domain \citep{kang2019cycle}:
\begin{align}
\begin{split}
\ell_{identity}(G,F) & =\mathbb{E}_{x \sim P_\Xc}[||G(x)-x||_1] + \mathbb{E}_{y \sim P_\Yc}[||F(y)-y||_1].
\label{eqn:cyclegan_identity_loss}
\end{split}
\end{align}

\begin{figure}[!t]
\centering
\includegraphics[width=0.8\linewidth]{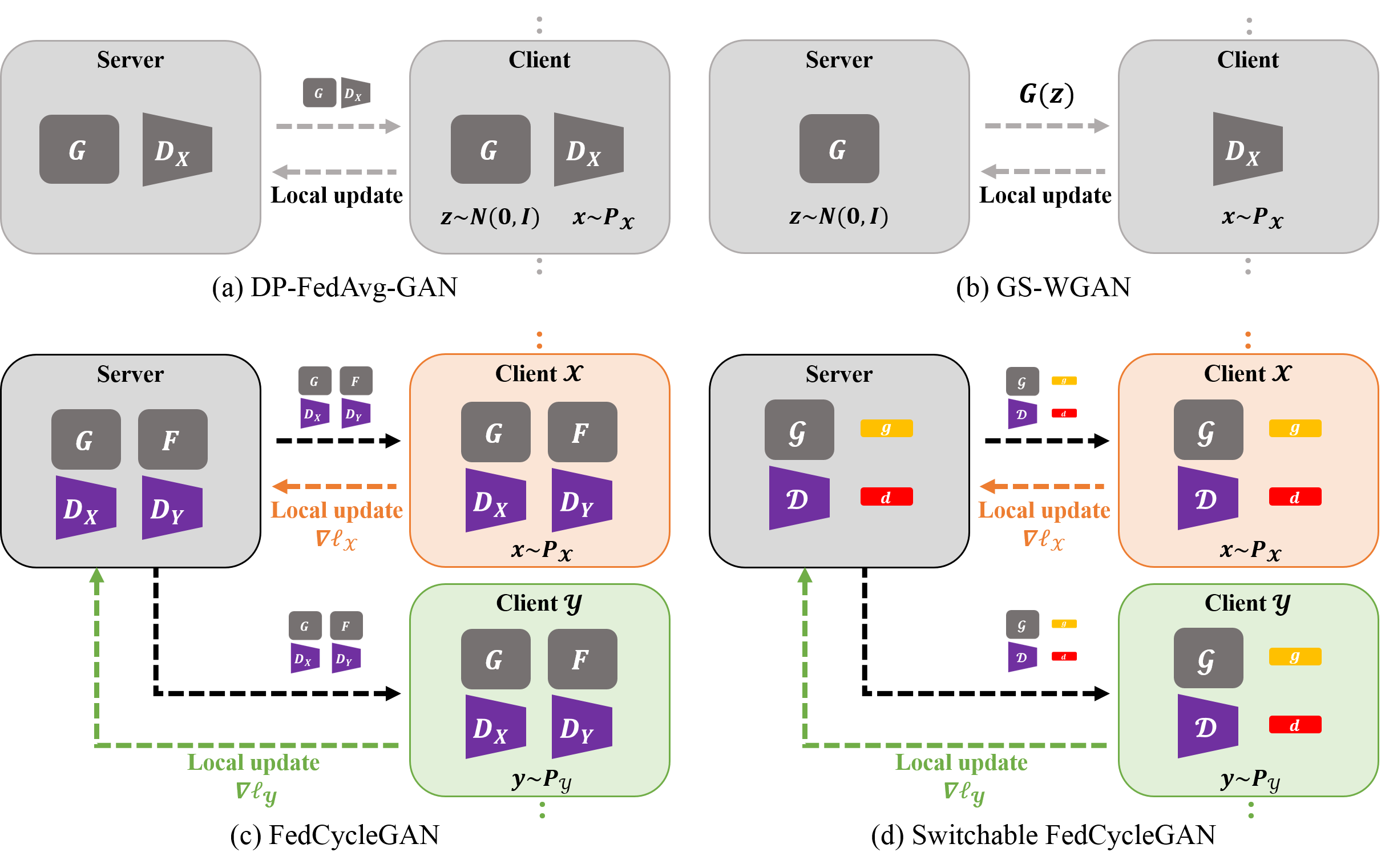}
\vspace{-0.2cm}
\caption{{Overview of (a) DP-FedAvg-GAN, (b) GS-WGAN, (c) FedCycleGAN, and (d) Switchable FedCycleGAN.}}
\label{fig:comp}
\end{figure}

\paragraph{Open problem}
Note that in  DP-FedAvg-GAN, clients have an access of both random noise $z\sim P_\Zc$ and the true image $x\sim P_\Xc$, of which
situation is different from our federated CycleGAN scenario, where none of the clients have an
access of  images from both $\Xc$ and $\Yc$ domains.
Although  GS-WGAN split the random noise $z\sim P_\Zc$ and the true image $x\sim P_\Xc$  to server and client, respectively,
the server only has a generator whereas the discriminator exists only at the client. This asymmetric architecture cannot be used  in CycleGAN
since every client has its own generator to translate to the other domain.
Therefore, CycleGAN  in a federated  scenario has been an open problem so far.

\section{Federated CycleGAN} \label{section Federated CycleGAN}
   
   \subsection{Standard form}
   As shown in Fig.~\ref{fig:comp}(c), to enable CycleGAN training in a federated setting, each client should have two generators ($G$ and $F$) and two  discriminators ($D_X$ and $D_Y$).
Then, the key question is whether gradient update is possible without accessing the other domain data.
Amazingly, this problem can be addressed with a simple observation.  Specifically, note that the CycleGAN loss in \eqref{eqn:cyclegan_overall_loss} can be decomposed into
domain specific two local objectives:
\begin{align}\label{eq:decomp}
\begin{split}
\ell_{CycleGAN}(G,F,D_X,D_Y) 
&= \ell_\Xc(G,F,D_X,D_Y)+\ell_\Yc(G,F,D_X,D_Y)
\end{split}
\end{align}
where $\ell_\Xc$ and $\ell_\Yc$ are local objectives that only use data in $\Xc$ and $\Yc$ domains, respectively:
\begin{align}
\ell_\Xc(G,F,D_X,D_Y)&=
\mathbb{E}_{x \sim P_\Xc}[\log D_X(x)]+ \mathbb{E}_{x \sim P_\Xc}[\log (1-D_Y(F(x)))]\notag\\&+{\lambda} \mathbb{E}_{x \sim P_\Xc}[||G(F(x))-x||_1]\label{eq:domainX}\\
\ell_\Yc(G,F,D_X,D_Y)&=\mathbb{E}_{y \sim P_\Yc}[\log (1-D_X(G(y)))]+\mathbb{E}_{y \sim P_\Yc}[\log D_Y(y)]\notag\\&+{\lambda} \mathbb{E}_{y \sim P_\Yc}[||F(G(y))-y||_1]\label{eq:domainY}
\end{align}
Accordingly, in our FedCycleGAN, instead of using the original CycleGAN loss $\ell_{CycleGAN}$,
the client with the data in $\Xc$ domain
uses $\ell_\Xc$, whereas the other client in $\Yc$ domain employs the loss $\ell_\Yc$.
Accordingly, clients  compute their gradients of their own losses and transmits those gradients to the server, after which the server train the global models by using the local gradients without accessing the data itself.
The exact local objective decomposition \eqref{eq:decomp} is indeed the key  that enables FedCycleGAN to retain the orignal CycleGAN performance.

\subsection{Beyond two clients}
Thanks to the use of domain specific loss functions in \eqref{eq:domainX} and \eqref{eq:domainY}, the extension to multiple clients
is in fact straight forward.
Specifically, as shown in {Fig.~\ref{fig:protocol}},
clients having images in $\Xc$ domain use $\ell_\Xc$ as their local training objective,
whereas the clients with $\Yc$ domain images employ $\ell_\Yc$ as their losses.
Then, the sever randomly selects $N$ clients either in $\Xc$ and $\Yc$ domain to receive their local gradients.
Local gradients from $N$ clients are averaged as in \citep{mcmahan2017communication} and {the server uses the averaged gradient to train networks.
Note that training FedCycleGAN from multiple clients can be viewed as training with a stochastic gradient using multiple mini-batches made up of specific domains.}
This training process is iterated until the specified number of epochs.
Thanks to this scalability, our federated CycleGAN enables training an unsupervised image-to-image translation models in a multi-client environment  without sharing privately sensitive data.
%
%
%

\subsection{Switchable form}
   
   Note that the local loss $\ell_\Xc$ and $\ell_\Yc$ are functions of two generators $G,F$ and two discriminators $D_X,D_Y$, so each
   client transmits the gradients of the four networks. Since the generators and discriminators are usually in complicated structures for the image-to-image translation tasks,
   the bandwidth requirement for the gradient transmission could be demanding.

   Recently, Gu et al \citep{gu2021adain} proposed a switchable CycleGAN in which a single generator can transfer an image in $\Yc$ to an output image in $\Xc$, and the generator can also be switched to a generator that translates an input in $\Xc$ into an output in $\Yc$ by changing the AdaIN code. 
   Inspired by this, 
   here we also propose a switchable FedCycleGAN where both generator and discriminator have switchable architecture {as shown in Fig.~\ref{fig:FedCycleGAN}(b)}.
   Specifically, 
   we apply different AdaIN code to a shared discriminator and a shared generator when training each client's model.
   
   Formally, our switchable generator $\Gc$ and the discriminator $\Dc$ can be defined as:
   \begin{eqnarray}
   \Gc(\cdot,c) = \begin{cases} G(\cdot), &\mbox{if~} c=g(0) \\ F(\cdot), &\mbox{if~} c=g(1)\end{cases},&&   \Dc(\cdot,c) = \begin{cases} D_X(\cdot), &\mbox{if~} c=d(0) \\ D_Y(\cdot), &\mbox{if~} c=d(1)\end{cases}
   \end{eqnarray}
   where  $g(n)$ and $d(n)$ denote the AdaIN code generators  for generator and discriminators, respectively, with a pre-defined input code index $n$.
 {Fig.~\ref{fig:architecture} (a) shows the architecture of the switchable generator, which is composed of convolution layers, Leaky ReLU, upsampling layers, and AdaIN layers with the AdaIN code generator consisting of fully connected layers. The AdaIN code generator produces mean and variance for each feature map in the generator, and the generator can be switchable by using these mean and variance vectors.
   The switchable discriminator consists of convolution layers, Leaky ReLU and AdaIN layers with the AdaIN code generator as shown in Fig.~\ref{fig:architecture} (b).}

\begin{figure}[!hbt]
\centering
\includegraphics[width=\linewidth]{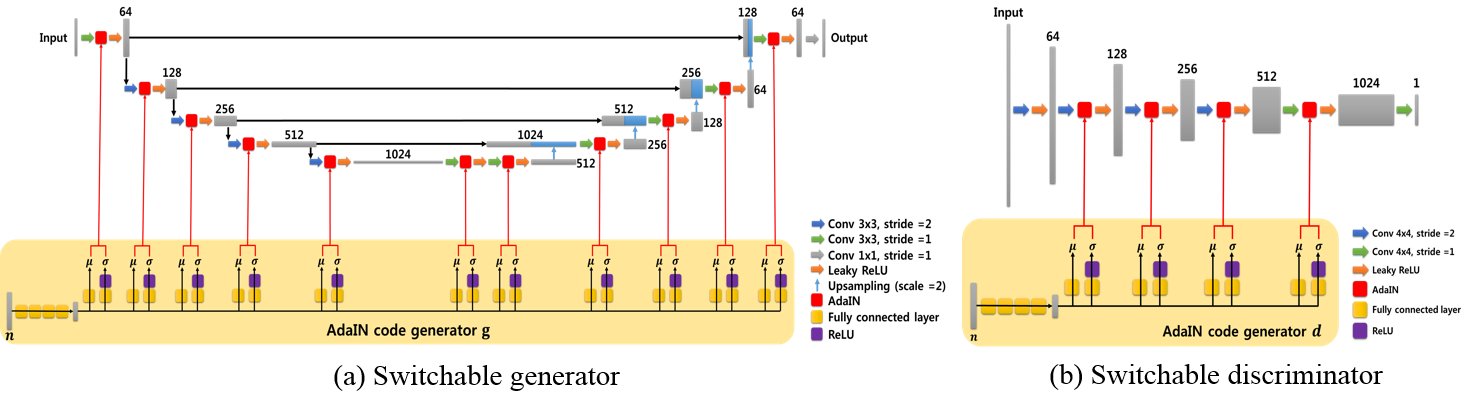}
\vspace{-0.5cm}
\caption{{Architecture of our (a) the switchable generator and (b) the switchable discriminator with the AdaIN code generators. The number on the top of the feature maps indicates the number of channels.}}
\label{fig:architecture}
\end{figure}

Using the switchable generator and discriminator, the local losses for $\Xc$ and $\Yc$ domains can be simplified as follows:
\begin{align*}
\ell_\Xc(\Gc,\Dc,g,d)&=
\mathbb{E}_{x}[\log \Dc(x,d(0))]+ \mathbb{E}_{x}[\log (1-\Dc(\Gc(x, g(1)), d(1)))] \notag\\
&~+{\lambda} \mathbb{E}_{x}[||\Gc(\Gc(x,g(1)),g(0))-x||_1]\notag\\
\ell_\Yc(\Gc,\Dc,g,d)&=\mathbb{E}_{y}[\log (1-\Dc(\Gc(y,g(0)), d(0)))]+\mathbb{E}_{y}[\log \Dc(y,d(1))]\notag\\
&~+{\lambda} \mathbb{E}_{y}[||\Gc(\Gc(y,g(0)),g(1))-y||_1]
\end{align*}

\newlength{\oldintextsep}
\setlength{\oldintextsep}{\intextsep}

\setlength\intextsep{0pt}
\begin{wraptable}{r}{0.5\textwidth}
  \caption{{Comparison of number of trainable parameters to transmit.}}
    \vspace*{0.3cm}
  \label{table1}
  \centering
  \scalebox{0.8}{
  \begin{tabular}{cccc}
  	\toprule
  	\multicolumn{2}{c}{CycleGAN} &\multicolumn{2}{c}{Ours (switchable)}	\\
    \cmidrule(r){1-2} \cmidrule(r){3-4}
    network 	    	& $\#$ of params  	&network 			& $\#$ of params				\\
    \midrule
    $G$					& 23,598,915  		& $\Gc$				& 23,598,915	 			\\
    $F$					& 23,598,915  		& $g$				& 544,896	 			\\
    $D_X$				& 11,162,561  		& $\Dc$				& 11,162,561	 			\\
    $D_Y$				& 11,162,561  		& $d$				& 270,336	 			\\
    \midrule
    Total				& 69,522,952  		& Total		& 35,576,708	 			\\
    \bottomrule
  \end{tabular}}
\end{wraptable}

Therefore, in contrast to the standard  form FedCycleGAN,
in the switchable form FedCycleGAN, each client can only transmit
the gradient with respect to the common generator and discriminator $\Gc, \Dc$ in addition to the
AdaIN code generators $g$ and $d$. The key point is that the code generators $g$ and $d$ are very
light, so the transmission bandwidth can be significantly reduced.
The number of parameters of networks for switchable form is approximately 35 million, while standard form requires approximately 69 million parameters in our experiments as in Table~\ref{table1}. 

Then, the training process of our switchable FedCycleGAN in each training round is described in Fig.~\ref{fig:protocol}. First,  the central server sends the current parameters $W^{(k)}_{Global}$ of the generator, the discriminator, and the AdaIN code generators to the selected clients. 
Then,   the selected clients compute their  local stochastic gradients $\nabla \ell_\Xc$ or $\nabla \ell_\Yc$ using fake samples and its own real samples by changing the AdaIN codes. Each client then sends their local gradients to the central server.  The central server updates the parameters of the generator, the discriminator, and the AdaIN code generators using local stochastic gradients from clients. This process is repeated until the networks converge. 

\section{Experimental results}

\subsection{Methods}
   To evaluate the performance of our method, we applied our proposed method to various style transfer tasks  \citep{zhu2017unpaired} and the low-dose computed tomography (CT) denoising task \cite{kang2019cycle}. We also compared our method with non-federated CycleGAN. 
{Specifically, we generate image-to-image translation results for each task by using three different methods: (1) {CycleGAN}, (2) {FedCycleGAN}, and (3) {switchable FedCycleGAN}. 
   For {CycleGAN}, we trained networks in a centralized setting by optimizing Eq. \eqref{eqn:cyclegan_min-max_overall_loss} using a generator and a discriminator, in which network architectures are basically same as Fig.~\ref{fig:architecture} except for the AdaIN layers (see Appendix).
   {FedCycleGAN} is trained by using gradient information without sharing data from two clients as described in Section~\ref{section Federated CycleGAN} {using the same generator and discriminator used in the CycleGAN}.
   In {Switchable FedCycleGAN}, a switchable generator and a switchable discriminator are used, and the training was carried out with local gradients from clients as described in Section~\ref{section Federated CycleGAN}.}
   To train {three different methods, we used the same training setting of $\lambda=10$ with identity loss where loss weight is $5$}, and used Adam optimizer \citep{kingma2014adam} with $\beta_1=0.5$ and $\beta_2=0.999$ for 200 epochs.  During the first 100 epochs, the learning rate was fixed at 0.0002 and then gradually reduced to 0 for the remaining 100 epochs. Pytorch \citep{paszke2019pytorch} and a NVIDIA GeForce RTX 3090 were utilized for the implementation.

\begin{table}[!b]
  \caption{{Comparison of various methods for image style transfer and low-dose CT denoising ($\downarrow$: lower is better, $\uparrow$: higher is better).}}
  \label{table2}
  \centering
  \scalebox{0.8}{
  \begin{tabular}{@{}lcccccccc@{}}
  	\toprule
  				&\multicolumn{6}{c}{Style transfer}	&\multicolumn{2}{c}{Low-dose CT denoising}	\\
    \cmidrule(r){2-7} \cmidrule(r){8-9}
    			&\multicolumn{2}{c}{summer-to-winter}	&\multicolumn{2}{c}{photo-to-vangogh}	&\multicolumn{2}{c}{horse-to-zebra} &\multirow{2}{*}{PSNR [dB] $\uparrow$} & \multirow{2}{*}{SSIM $\uparrow$} \\
    \cmidrule(r){2-7}
    			&FID $\downarrow$	&IS	$\uparrow$	&FID $\downarrow$	&IS	$\uparrow$	&FID $\downarrow$	&IS	$\uparrow$	\\
    \midrule
  Input					&$\cdot$ 		&$\cdot$		&$\cdot$			&$\cdot$		&$\cdot$			&$\cdot$									&32.5132 			&0.7411 \\
  CycleGAN     			&75.40			&\textbf{2.71}	&119.85				&3.74		 	&124.45				&1.90 										&37.1173			&0.8654   	\\
  \textbf{Ours}  				&\textbf{74.72} &2.64  			&120.24				&3.53			&\textbf{107.17}	&1.70										&\textbf{37.1478}	&\textbf{0.8656}	\\
  \textbf{Ours (switchable)}	&76.65  		&2.57			&\textbf{118.35}	&\textbf{4.28}	&121.90				&\textbf{1.97}								&37.0723       		& 0.8637	\\
    \bottomrule
  \end{tabular}}
\end{table}

\paragraph{Style transfer tasks}
{ We conducted various style transfer tasks such as summer to winter, photo to Van Gogh, and horse to zebra by using publicly available datasets from \citep{zhu2017unpaired}.
   We assume two clients, and each client has data from other classes for a federated learning scenario.}
   The    summer to winter dataset consists of unpaired training sets (1231 summer images and 962 winter images) and test sets (309 summer images, 238 winter images).   The  photo to {Van Gogh} dataset is composed of unmatched training images (6287 photo images and 400 Van Gogh images) and test images (751 photo images and 400 Van Gogh images).
The      horse to zebra dataset is divided by training set (1067 horse and 1334 zerbra) and test images (120 horse and 140 zebra). Images from each class are unpaired.
   To train style transfer tasks such as summer-to-winter, {photo-to-Van Gogh}, and horse-to-zebra, we augmented images by resizing them to 286 $\times$ 286 pixels and then cropping them to get patches with the size of 256 $\times$ 256 pixels. We also applied a random horizontal flip to images with a probability of 0.5. The batch size was set to 4.

\paragraph{Low-dose CT denoising}
{X-ray computed tomography (CT) is one of the most important imaging systems for clinical use. However, the radiation dose exposed by CT scan increases the cancer risk for patients. A low-dose CT scan reduces the risk of radiation, while a high level of noise can be an obstacle to clinical diagnosis, which is why low-dose CT denoising is required.
   We utilized AAPM CT dataset  used in \citep{kang2017deep, kang2018deep} consisting of projection data from the AAPM 2016 Low Dose CT Grand Challenge \citep{mccollough2017low}. All data were completely anonymized.  
   For the unsupervised image-to-image translation, we trained our method with unpaired image sets. One client have low-dose CT image while another have routine-dose CT image.}
The size of CT image slices is 512 $\times$ 512 pixels. In the AAPM low-dose CT challenge, to simulate low-dose CT, Poisson noise was added into the projection data which corresponds to 25 $\%$ of the full dose CT.
   We used 8 patients data composed of the routine dose and low dose images for training set, and one patient data is used for the test set. The number of training slices is 3236 while 350 slices are used for the test.

In our two clients experiment,
   one client have 3236 low-dose CT images while another have 3236 routine-dose CT images from the AAPM CT dataset. We randomly shuffle the order of data in each training round to make an unpaired image set.
   To train the low-dose CT denoising, we cropped the input image with the size of 512 $\times$ 512 pixels to get samples with the size of 128 $\times$ 128 pixels. We also flipped images horizontally and vertically for data  augmentation. The batch size was set to 8.
   For multi-clients experiment,
   we consider a scenario where four clients participate the training procedure. Routine-dose data from 8 patients are divided and given two clients (1948 and 1646 slices for each clients), while low-dose CT data from 8 patients are divided and given two other clients (1948 and 1646 slices for each clients). The order of data is shuffled. Other training setting is same as that of {the two clients experiment for low-dose CT denoising.}

\begin{figure}[!t]
\centering
\includegraphics[width=0.8\linewidth]{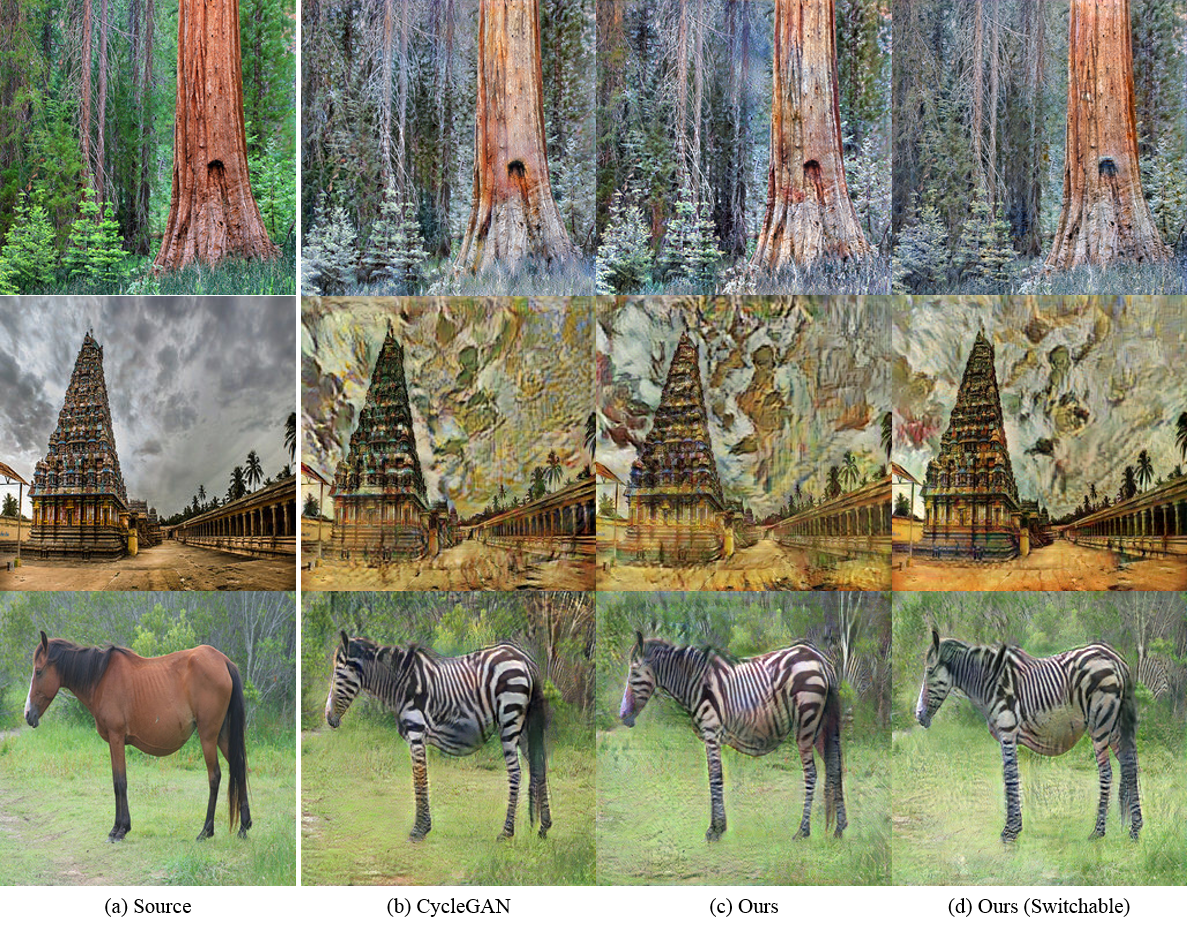}
\vspace{-0.3cm}
\caption{{Image style transfer results by various methods for summer-to-winter (first row), photo-to-Van Gogh (second row), and horse-to-zebra (third row).}}
\label{fig:results_style}
\end{figure}


\subsection{Style transfer results}

   Fig. \ref{fig:results_style} shows results of various style transfer tasks. {For all tasks, our federated learning frameworks (FedCycleGAN and Switchable FedCycleGAN) successfully transfer source images to the target domain, which is comparable to the results of non-federated CycleGAN.}
   For the quantitative comparison of sample qualities, we calculated Inception Score (IS) \citep{NIPS2016_8a3363ab} and Frechet Inception Distance (FID) \citep{NIPS2017_8a1d6947}. Table~\ref{table2} shows IS and FID values from three different methods for various style transfer tasks.
   In summer to winter style transfer task, FedCycleGAN achieves the best FID score.
   In photo to Van Gogh style transfer task, Switchable FedCycleGAN achieves the best FID and IS compared to other methods.
   In horse to zebra style transfer task, FedCycleGAN achieves the best FID score and Switchable FedCycleGAN have better FID and IS than CycleGAN which is the non-federated baseline.
   In summary, our federated methods can achieve comparable and even better performance compared to the non-federated baseline without sharing data  in unsupervised style transfer tasks.

\subsection{Low-dose CT denoising}

\begin{figure}[!t]
\centering
\includegraphics[width=\linewidth]{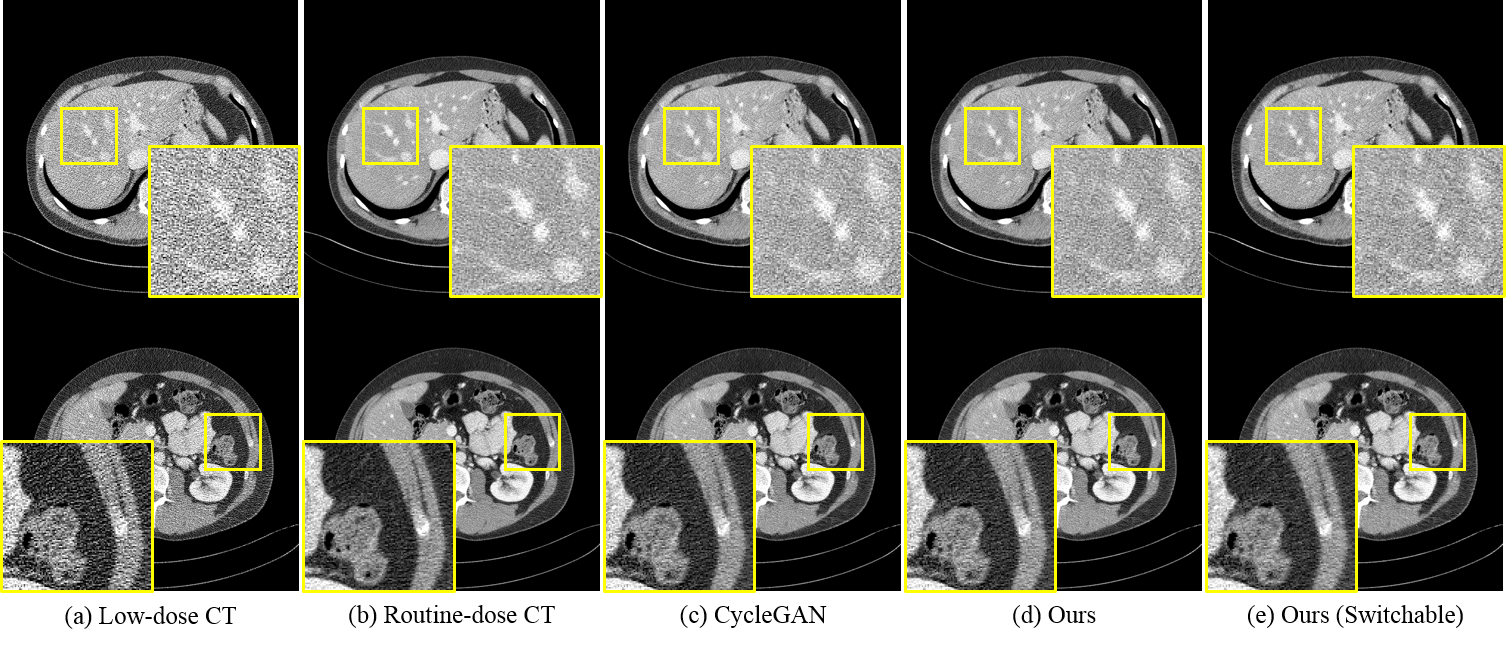}
\vspace{-0.6cm}
\caption{{Comparison of various methods for low-dose CT denoising. Intensity range of images is (-160, 240) [HU] (Hounsfield Unit).}}
\label{fig:results_ct}
\end{figure}


  To evaluate the performance of denoising results, we calculated the peak signal-to-noise ratio (PSNR) and the structural similarity index metric (SSIM) \citep{wang2004image} values. Table~\ref{table2} also lists the evaluation metrics for low-dose CT denoising results on AAPM CT dataset.
Notice that FedCycleGAN achieves the highest PSNR and SSIM values. 
  We found that results of non-federated CycleGAN and results of Switchable FedCycleGAN showed small differences in terms of PSNR and SSIM. The difference in PSNR values is 0.045 dB, while the difference in SSIM values is 0.0017.
  {Fig. \ref{fig:results_ct} shows denoising results from various methods. We found that our federated methods produce successful denoising results, which are similar to the routine dose target domain images. The results are also  comparable to those of non-federated CycleGAN.}

\paragraph{Multiple clients experimental results}
We also conducted the multi-clients federated CycleGAN experiment in which four clients participate using Switchable FedCycleGAN  architecture.
Here,  two clients have non-overlapping routine-dose CT images while two other clients have different low-dose CT images. At each training round, we randomly select $N$ clients to train Switchable FedCycleGAN, in which $N$ is the constant number.
   Table~\ref{table3} lists quantitative comparison with various $N$ values. Note that as $N$ increases, the performance increases as the number of accessible mini-batch increases. When we use $N = 4$, the PSNR value was comparable to that of Switchable FedCycleGAN in Table~\ref{table2} and achieves better SSIM value compared to other methods in Table~\ref{table2}.
\begin{table}
  \caption{{Quantitative comparison of switchable FedCycleGAN for  low-dose CT denoising using multiple clients. At each round, the server randomly selects $N$ clients from four clients to receive local gradients.}}
  \label{table3}
  \centering
  \scalebox{0.9}{
  \begin{tabular}{lccccc}
    \toprule
     	     					& Input  	& N = 1		& N = 2		& N = 3		& N = 4		\\
    \midrule
    PSNR [dB] $\uparrow$		& 32.5132  	& 36.7815	& 36.8054	& 36.9034   & \textbf{37.0577} 			\\
    SSIM 	  $\downarrow$	   	& 0.7411 	& 0.8593    & 0.8611 	& 0.8646	& \textbf{0.8682}			\\
    \bottomrule
  \end{tabular}}
\end{table}
\section{Conclusions}

In this paper, we propose a federated CycleGAN and a switchable FedCycleGAN for privacy-preserving image-to-image translation.
   With our framework, a server does not require any private local data and needs only local gradients from clients. Our experimental results demonstrate that our method can be successfully applied to various unsupervised image translation tasks and show promising results compared to the non-federated counterpart.
This was possible thanks to the  exact local objective decomposition,  which could be extended to other multi-domain federated translation tasks \cite{choi2020stargan}.
   We believe that our framework gives a new direction for the study of federated and unsupervised image-to-image translation to real-world situations.

\paragraph{Limitation and negative societal impacts}
  {
   Although our framework does not require local data from clients, the gradient information is necessary to train image translation model. However, if gradient information is known, there is a possibility that hidden images of clients can be reconstructed as studied in \citep{NEURIPS2020_c4ede56b}.
   This means that the gradient information can be stolen for criminal purposes while our frameworks train the networks.  
   To ensure privacy of gradient information, techniques for privacy guarantee (e.g. differential privacy \citep{dwork2014algorithmic}) need to be applied in the future work.}

%
%

\renewcommand{\bibname}{References}
\renewcommand*{\bibfont}{\small}
\bibliographystyle{abbrvnat}


\appendix

\section{Appendix}

\subsection{Algorithm 1: Switchable FedCycleGAN}

Algorithm~\ref{algorithm:swfedcyclegan} describes the procedure of training a swtichable FedCycleGAN. 
First, a central server initializes parameters of switchable generator $\Gc$, switchable discriminator $\Dc$, and AdaIN code generators ($g$ and $d$). 
For each training round $k$, each client calculates a local gradient using its own data (in $\Xc$ or $\Yc$) and current parameters of networks from the server ($\theta_\Gc^{(k-1)}$, $\theta_\Dc^{(k-1)}$, $\theta_g^{(k-1)}$, $\theta_d^{(k-1)}$), and then gradient information is transmitted to the server. 
\textsc{GetLocalGradient} in Algorithm~\ref{algorithm:swfedcyclegan} describes how to calculate a local gradient in each client. Each client computes a domain specific local objective ($\ell_{\Xc}$ or $\ell_{\Yc}$) using a batch from its own data. Then, a stochastic gradient for each network can be calculated and transmitted to the server.
After clients transmit their local gradient information to the server, the server sums the gradients from clients and updates parameters of networks using the summed gradients by a gradient descent with learning rate $\eta$ . The training round is repeated up to the total number of rounds (epochs) $K$.

\subsection{Network Architecture}

   Fig.~\ref{fig:generator_baseline} shows the architecture of the generator for CycleGAN and FedCycleGAN models. The generator is based on U-net \citep{RFB15a} structure and consists of several convolution layers, Leaky ReLU, upsampling layers, and instance normalization layers. For upsampling layers, we used nearest neighbour upsampling. Note that we use the same architecture in Fig.~\ref{fig:generator_baseline} for style transfer tasks, while we added a skip connection between the input and output layer in the low-dose CT denoising task for residual learning that leads to better performance on denoising problems \citep{zhang2017beyond}.
   
   The architecture of the discriminator for CycleGAN and FedCycleGAN models is shown in Fig.~\ref{fig:discriminator_baseline}. It consists of several convolution layers, Leaky ReLU, and instance normalization layers.

\subsection{Convergence of GAN loss}
Similar to the vanilla GAN loss \citep{goodfellow2014generative}, the LSGAN \citep{mao2017least} can also be decomposed into domain-specific local objectives so 
  we used LSGAN loss for all experiments to stabilize training.
  Fig~\ref{fig:gan_loss_plot} shows GAN losses during training of a low-dose CT denoising task. 
Similar to the centralized traininng,
our federated cycleGAN training converges to the optimal state where the LSGAN loss is 0.25.

\subsection{Additional results}
  We present additional results for style transfer tasks and low-dose CT denoising task.
  Fig~\ref{fig:appendix_summter2winter} shows image translation results for summer-to-winter.
  Fig~\ref{fig:appendix_photo2vangogh} shows image style transfer results for photo-to-Van Gogh.
  Fig~\ref{fig:appendix_ct_denoising} shows denoising results for low-dose CT images.
  Note that our methods produce comparable results compared to non-federated CycleGAN.


\begin{algorithm}[!hbt]
\SetAlgoLined
\SetKwInOut{Input}{Input}\SetKwInOut{Output}{Output}
\SetKwFunction{GetLocalGradient}{GetLocalGradient}
\SetKwProg{Fn}{}{:}{}

 \Input{total number of rounds (epochs) $K$, learning rate $\eta$}
 \Output{switchable generastor $\Gc$, switchable discriminator $\Dc$, AdaIN code generators $g$ and $d$}
 \BlankLine
 Initialize $\theta_\Gc$, $\theta_\Dc$, $\theta_g$, $\theta_d$ in server\;
 \For{each round k = 1, 2, ..., $K$}{
  \For{each client i \textbf{in} $\{\Xc, \Yc\}$} {
  	${g_{\theta_\Gc}}^{(i)}, {g_{\theta_\Dc}}^{(i)}, {g_{\theta_g}}^{(i)}, {g_{\theta_d}}^{(i)} \leftarrow$ \GetLocalGradient(i, $\theta_\Gc^{(k-1)}$, $\theta_\Dc^{(k-1)}$, $\theta_g^{(k-1)}$, $\theta_d^{(k-1)}$)\;
  }
  	${\theta_\Gc}^{(k)} \leftarrow {\theta_\Gc}^{(k-1)} - \eta \sum_{i} {g_{\theta_\Gc}}^{(i)} $ \;
  	${\theta_\Dc}^{(k)} \leftarrow {\theta_\Dc}^{(k-1)} - \eta \sum_{i} {g_{\theta_\Dc}}^{(i)} $ \;
  	${\theta_g}^{(k)} \leftarrow {\theta_g}^{(k-1)} - \eta \sum_{i} {g_{\theta_g}}^{(i)} $ \;
  	${\theta_d}^{(k)} \leftarrow {\theta_d}^{(k-1)} - \eta \sum_{i} {g_{\theta_d}}^{(i)} $ \;
 }
 \KwRet{$\Gc$, $\Dc$, $g$, $d$}
 
 \BlankLine
 
 \Fn{\GetLocalGradient(i, $\theta_\Gc$, $\theta_\Dc$, $\theta_g$, $\theta_d$)}{
 	$b \leftarrow$ (sample batch from client $i$) \;
	${g_{\theta_\Gc}}^{(i)} \leftarrow \nabla_{\theta_\Gc} \ell_{i}(\Gc, \Dc, g, d; b)$ \;
	${g_{\theta_\Dc}}^{(i)} \leftarrow \nabla_{\theta_\Dc} \ell_{i}(\Gc, \Dc, g, d; b)$ \;
	${g_{\theta_g}}^{(i)} \leftarrow \nabla_{\theta_g} \ell_{i}(\Gc, \Dc, g, d; b)$ \;
	${g_{\theta_d}}^{(i)} \leftarrow \nabla_{\theta_d} \ell_{i}(\Gc, \Dc, g, d; b)$ \;
	\KwRet{${g_{\theta_\Gc}}^{(i)}, {g_{\theta_\Dc}}^{(i)}, {g_{\theta_g}}^{(i)}, {g_{\theta_d}}^{(i)}$}
 } 
 
 \caption{Switchable FedCycleGAN}
 \label{algorithm:swfedcyclegan} 
\end{algorithm}

\begin{figure}[!ht]
\centering
\includegraphics[width=1.0\linewidth]{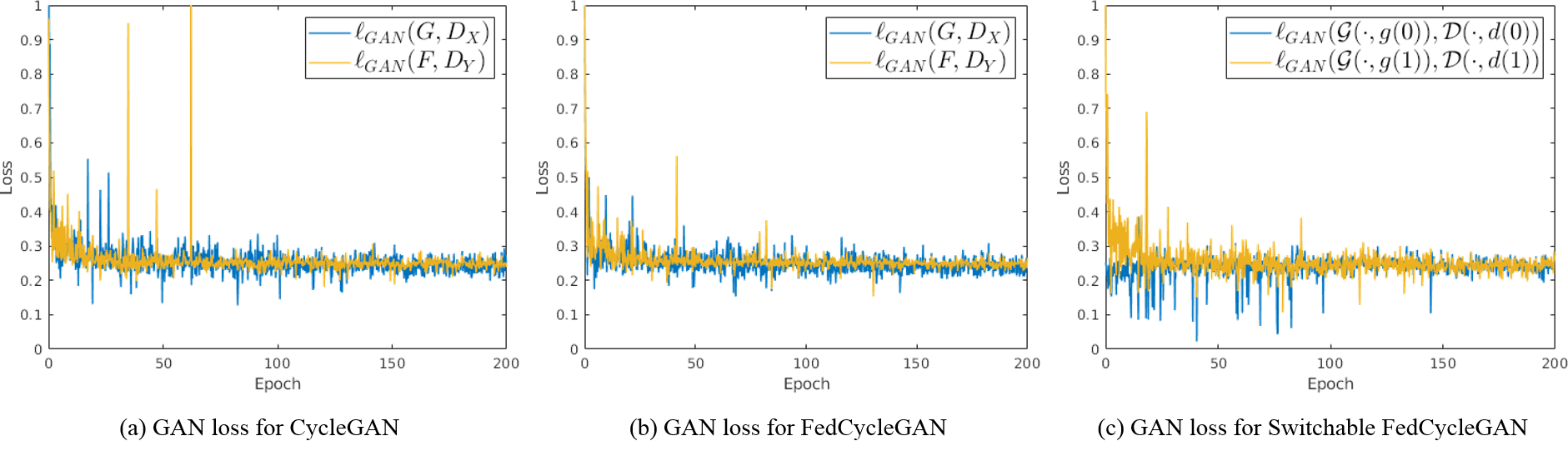}
\vspace{-0.5cm}
\caption{Convergence of GAN loss during training of a low-dose CT denoising task.}
\label{fig:gan_loss_plot}
\end{figure}

\begin{figure}[!ht]
\centering
\includegraphics[width=1\linewidth]{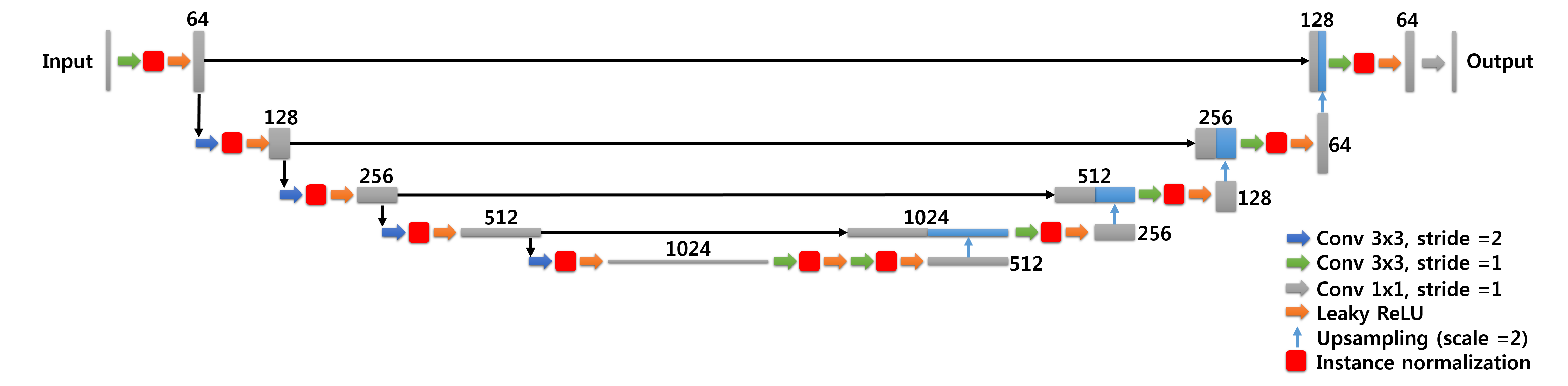}
\vspace{-0.1cm}
\caption{Architecture of the generator for CycleGAN and FedCycleGAN models.}
\label{fig:generator_baseline}
\end{figure}

\begin{figure}[!ht]
\centering
\includegraphics[width=0.7\linewidth]{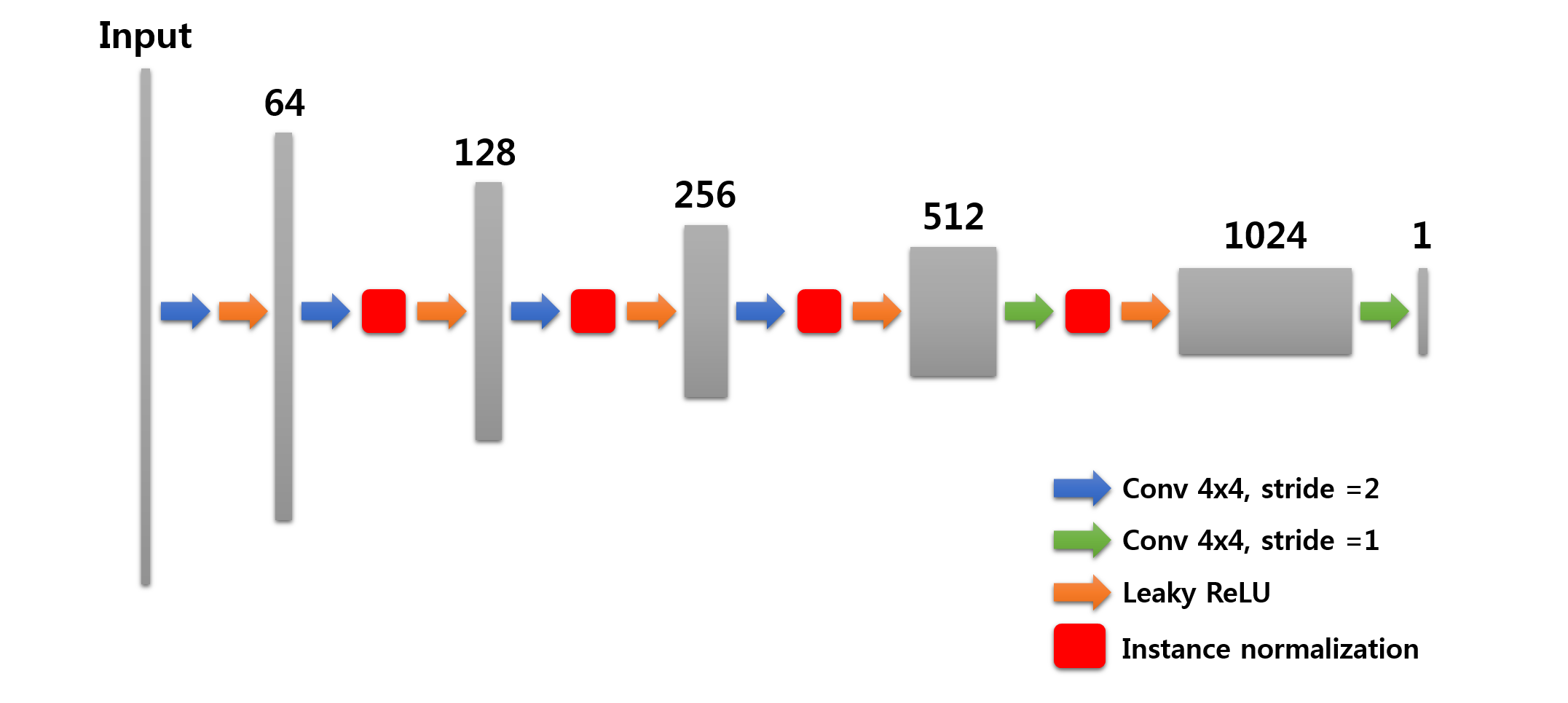}
\vspace{-0.1cm}
\caption{Architecture of the discriminator for CycleGAN and FedCycleGAN models.}
\label{fig:discriminator_baseline}
\end{figure}

\begin{figure}[]
\centering
\includegraphics[width=1.0\linewidth]{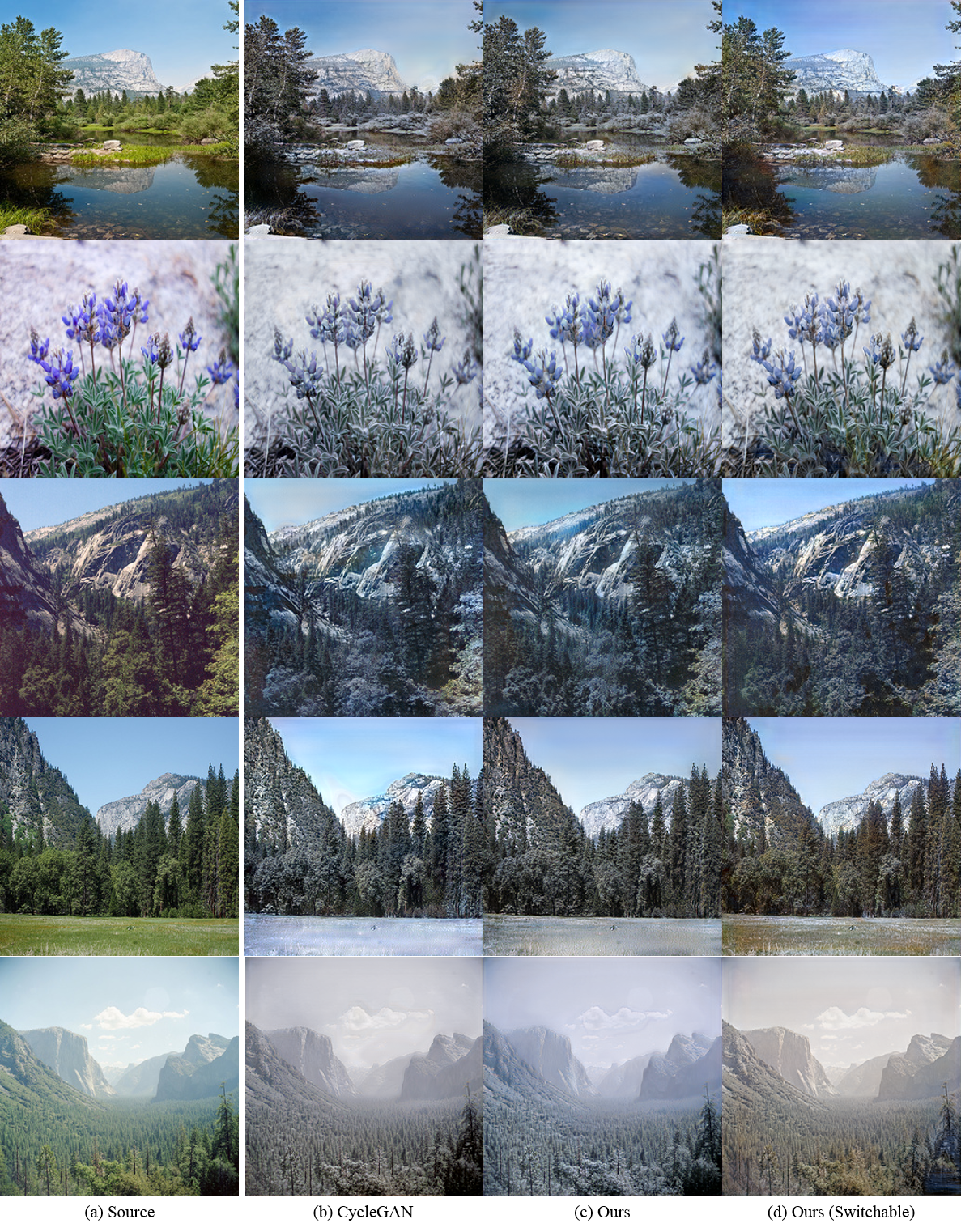}
\vspace{-0.5cm}
\caption{Image style transfer results for summer-to-winter task.}
\label{fig:appendix_summter2winter}
\end{figure}

\begin{figure}[]
\centering
\includegraphics[width=1.0\linewidth]{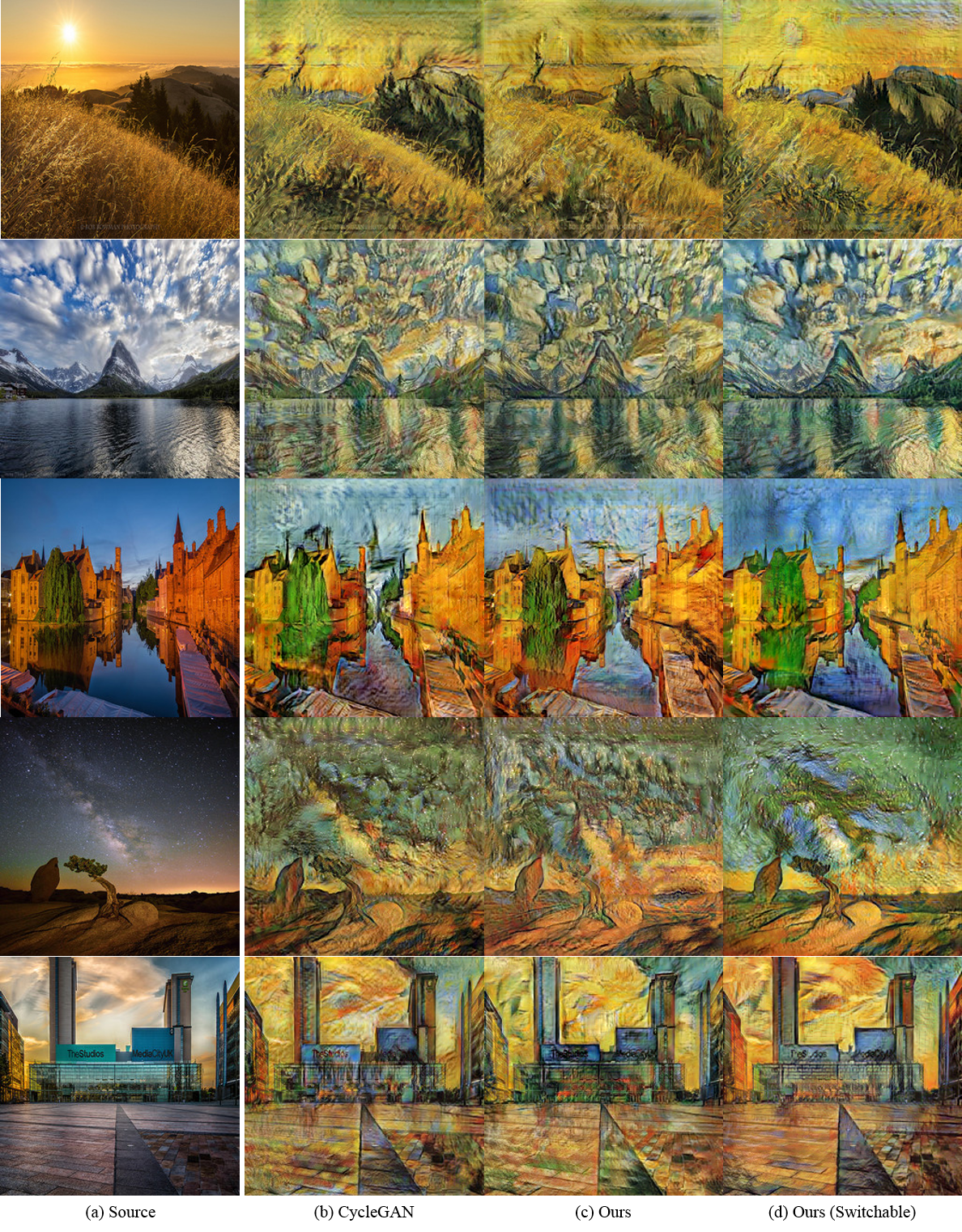}
\vspace{-0.5cm}
\caption{Image style transfer results for photo-to-Van Gogh task.}
\label{fig:appendix_photo2vangogh}
\end{figure}

\begin{figure}[]
\centering
\includegraphics[width=1.0\linewidth]{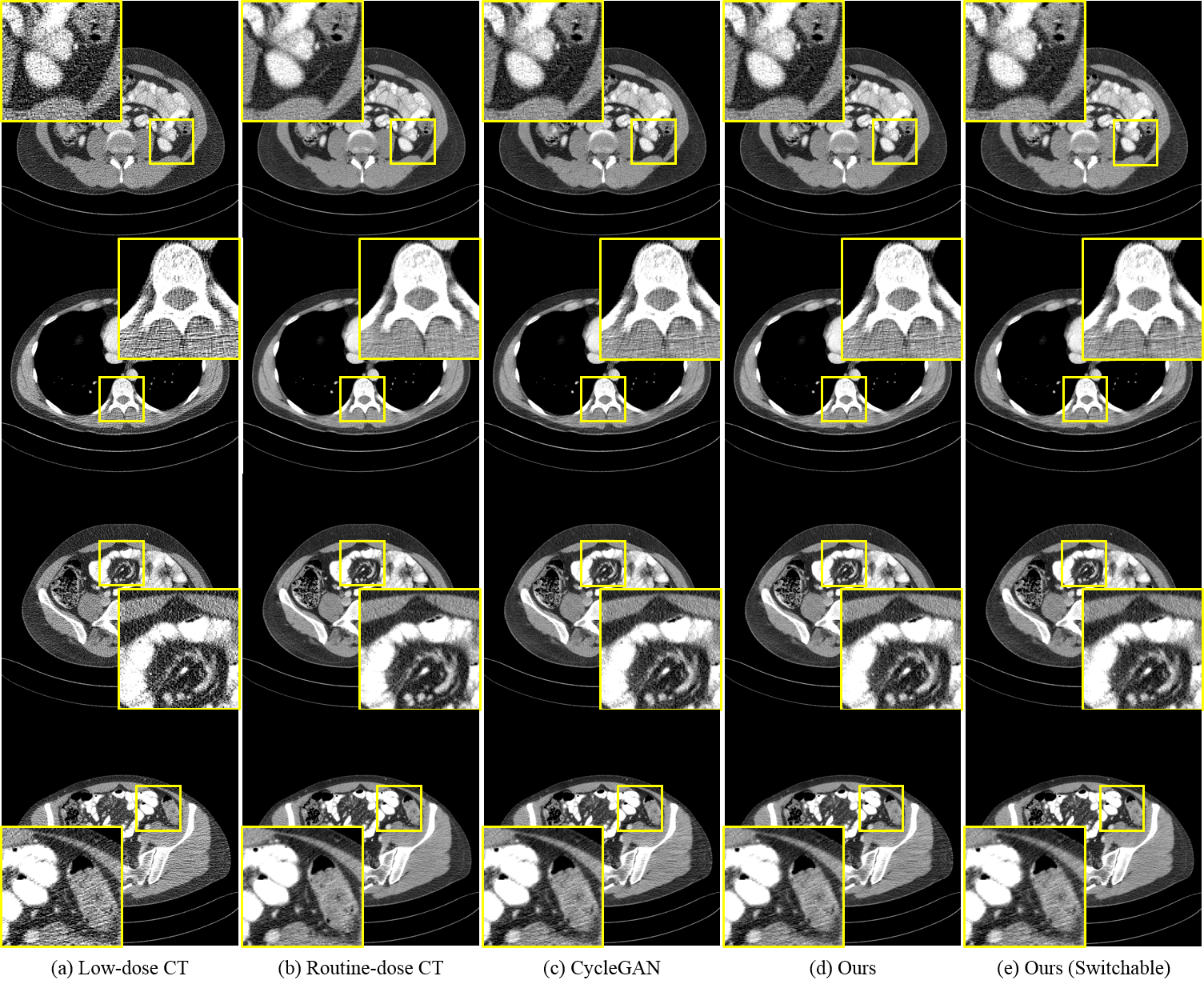}
\vspace{-0.5cm}
\caption{Results for low-dose CT denoising task. Intensity range of images is (-160, 240) [HU] (Hounsfield unit)}
\label{fig:appendix_ct_denoising}
\end{figure}

\end{document}